\documentclass[10pt,twocolumn,letterpaper]{article}

\usepackage{cvpr}              

\usepackage[normalem]{ulem}
\usepackage{tabularray}
\usepackage{graphicx}  
\usepackage{subcaption}  
\usepackage{pifont}
\usepackage[accsupp]{axessibility}
\definecolor{darkgreen}{HTML}{00BB00}
\definecolor{darkred}{HTML}{DD0000}
\usepackage{multicol}
\usepackage{caption}

\definecolor{cvprblue}{rgb}{0.21,0.49,0.74}
\usepackage[pagebackref,breaklinks,colorlinks,allcolors=cvprblue]{hyperref}

\def\paperID{****} 
\def\confName{CVPR}
\def\confYear{2026}

\title{SECOS: Semantic Capture for Rigorous Classification in Open-World Semi-Supervised Learning\thanks{Accepted by CVPR 2026}}

\author{Hezhao Liu$^1$ \quad Jiacheng Yang$^1$ \quad Junlong Gao$^1$ \\ Mengke Li$^2$ \quad Yiqun Zhang$^3$ \quad Shreyank N Gowda$^4$ \quad Yang Lu$^{1 }$\thanks{Corresponding Author: Yang Lu (luyang@xmu.edu.cn)}\\
{\footnotesize $^1$Key Laboratory of Multimedia Trusted Perception and Efficient Computing, Ministry of Education of China, Xiamen University, Xiamen, China} \\ {\footnotesize $^2$College of Computer Science and Software Engineering, Shenzhen University, Shenzhen, China} \\ {\footnotesize $^3$School of Computer Science and Technology, Guangdong University of Technology, Guangzhou, China} \\ {\footnotesize $^4$School of Computer Science, University of Nottingham, Nottingham, UK}\\
{\tt\footnotesize 23020231154209@stu.xmu.edu.cn \quad 23020250157840@stu.xmu.edu.cn \quad jlgao@xmu.edu.cn} \\ {\tt\footnotesize mengkeli@szu.edu.cn \quad yqzhang@gdut.edu.cn \quad shreyank.narayanagowda@nottingham.ac.uk \quad luyang@xmu.edu.cn}
}

\begin{document}
\maketitle
\begin{abstract}
In open-world semi-supervised learning (OWSSL), a model learns from labeled data and unlabeled data containing both known and novel classes. In practical OWSSL applications, models are expected to perform rigorous classification by directly selecting the most semantically relevant label from a candidate set for each sample. Existing OWSSL methods fail to achieve this because novel samples are trained without explicit supervision, and these methods lack mechanisms to extract latent semantic information, resulting in predicted labels that have no semantic correspondence to candidate textual labels. To address this, we introduce SEmantic Capture for Open-world Semi-supervised learning (SECOS), which directly predicts textual labels from the candidate set without post-processing, meeting the requirements of practical OWSSL applications. SECOS leverages external knowledge to extract and align semantic representations across modalities for both known and novel classes, providing explicit supervisory signals for training novel classes. Extensive experiments demonstrate that even when existing OWSSL methods are evaluated under the more lenient post-hoc matching setting, SECOS still surpasses them by up to 5.4\% without such assistance, highlighting its superior effectiveness. Code is available at \href{https://github.com/ganchi-huanggua/OSSL-Classification}{https://github.com/ganchi-huanggua/OSSL-Classification}.
\end{abstract}    
\vspace{-15pt}
\section{Introduction}
\begin{figure}
\centering
{\includegraphics[width=\columnwidth]{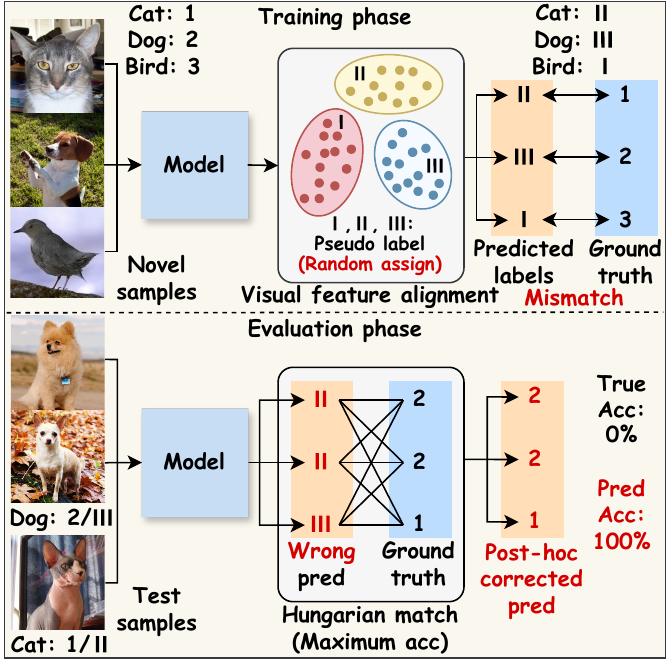}}
\vspace{-15pt}
\caption{Illustration of the limitations of existing OWSSL methods. During training, they overemphasize visual features and neglect latent semantic information in novel samples. During evaluation, they rely on Hungarian matching to align predicted labels with ground truth, highlighting that these methods perform clustering rather than RC-OWSSL.}
\label{fig:moti}
\vskip -20pt
\end{figure}

Open-world semi-supervised learning (OWSSL) \cite{cao2022open}, also known as generalized class discovery (GCD) \cite{vaze2022generalized}, extends standard semi-supervised learning (SSL) \cite{liu2025fate, liu2025mind, Cheng2025CGMatch} to settings where unlabeled data contain both known and novel classes, requiring algorithms to identify and organize novel samples into semantically coherent groups. However, existing OWSSL algorithms still fail to perform rigorous \textbf{classification} because they cannot directly predict textual labels, and the labels they output lack explicit semantic correspondence \cite{zheng2024textual, wen2023parametric, xiao2024targeted, wang2023discover}. A rigorous OWSSL classification task, which we term Rigorous Classification in OWSSL (RC-OWSSL), requires the model to directly select the most semantically relevant label for each sample from a candidate label set. This goal is consistent with standard classification tasks \cite{krizhevsky2012imagenet, he2016deep, simonyan2014very, szegedy2015going}. The candidate set includes labels from both known and novel classes, and the predicted label must reflect the true semantic information of the sample rather than being derived from post-hoc processing. 

Existing OWSSL methods fail to achieve RC-OWSSL because they provide limited explicit semantic supervision for training novel class samples and lack a mechanism to uncover the latent semantic information inherent in novel classes. As shown in \cref{fig:moti}, during training, these methods focus excessively on the visual features of samples while completely ignoring the latent semantic information embedded in novel samples \cite{xiao2024targeted, wang2023discover, zheng2024textual, liu2023open}. During evaluation, nearly all algorithms rely on Hungarian matching to force a post-hoc alignment between predicted pseudo labels and ground truth \cite{xiao2024targeted, wang2023discover, niu2024owmatch, zheng2024textual, fan2025learning, wen2023parametric}, which is highly unstable and can produce artificially high accuracy even when a substantial number of batch samples are misclassified. Moreover, in real-world scenarios, such a labeled test set is typically unavailable, making these methods impractical in realistic deployment. This underscores their inability to directly predict the textual labels, indicating that existing approaches do not perform RC-OWSSL but instead solve an OWSSL \textbf{clustering} task \cite{niu2024owmatch, zheng2024textual}. To motivate the design of our proposed framework for RC-OWSSL, we pose the following research questions.

\begin{enumerate}
    \item \textit{\textbf{What is the core ingredient missing in existing OWSSL methods that is essential for achieving RC-OWSSL?}} \\
    The core missing ingredient in prior work is semantic grounding, i.e., the ability to associate the visual features of novel classes with their linguistic meaning. Without capturing latent semantic information from novel classes, models cannot directly predict textual labels and meet the requirements of RC-OWSSL. 
    
    \item \textit{\textbf{What is the key challenge in capturing semantic information from unlabeled novel classes in RC-OWSSL, and how can it be addressed?}} \\
    The key challenge lies in the \emph{complete absence} of explicit supervision for novel classes, forcing the model to utilize external priors to uncover and reconstruct the missing correspondence between visual structures and semantic meanings. To address this, models with visual-textual \cite{radford2021learning} priors must be leveraged to compute semantic similarity between novel samples and candidate labels, aligning the two modalities and enabling effective semantic capture.

\end{enumerate}


Based on these questions, we propose SEmantic Capture for Open-world SSL (SECOS), a framework specifically designed for RC-OWSSL. SECOS enables direct prediction of candidate textual labels for novel class samples without relying on post-hoc alignment, making it applicable to real-world scenarios where labeled test sets are unavailable. SECOS leverages external knowledge to extract semantic representations from both known and novel class samples and aligns them with candidate textual labels. SECOS performs semantic extraction and alignment through three modules. Novel Class Semantic Compensation builds global semantic representations to compensate for the lack of supervision in novel classes. Batch-Wise Semantic Recapture refines sample-level semantics for fine-grained exploitation. Adapter for Semantic Feature Alignment aligns intermediate features with the generated supervisory signals, establishing the correspondence between visual structure and semantic meaning. Extensive experiments on generic and fine-grained datasets demonstrate that SECOS significantly outperforms existing OWSSL and GCD methods relying on Hungarian matching during testing. Our contributions can be summarized as follows.
\begin{itemize}
    \item We identify key principles for RC-OWSSL, which directly predict textual labels for novel samples without post-processing, revealing an intrinsic deviation between the goal of RC-OWSSL and existing clustering methods.
    \item We propose SECOS, a framework tailored for RC-OWSSL. SECOS leverages external knowledge to extract semantic representations from novel class samples, enabling direct classification in real-world scenarios.
    \item SECOS effectively achieves the RC-OWSSL challenge, demonstrated by up to 5.4\% improvement over state-of-the-art (SOTA) across seven benchmarks, even though existing approaches evaluate using Hungarian matching.
\end{itemize}

\section{Related Work}
\subsection{Open-World Semi-Supervised Learning}
OWSSL \cite{cao2022open} or GCD \cite{vaze2022generalized} aims to learn from labeled data of known classes and unlabeled data that includes both known and novel classes, reflecting more realistic learning scenarios than traditional SSL. Early OWSSL approaches typically follow a train-from-scratch paradigm, employing clustering-based \cite{xiao2024targeted, wang2023discover, liu2023open} or pseudo-labeling strategies \cite{niu2024owmatch, rizve2022openldn, rizve2022towards, guo2022robust} to utilize novel class samples. These methods extract visual representations and encourage feature separability through various unsupervised objectives. More recent works leverage pre-trained models, such as DINO \cite{vaze2022generalized, zhang2023promptcal, pu2023dynamic, wen2023parametric} and CLIP \cite{zheng2024textual, fan2025learning, ouldnoughi2023clip, wang2025get}, to incorporate prior knowledge, thereby enhancing feature representations and facilitating more effective downstream OWSSL. 

\subsection{Adapter-Based Fine-Tuning}
Adapter \cite{houlsby2019parameter} is a lightweight fine-tuning method that inserts small learnable modules into deeper layers of a frozen pre-trained model to adjust intermediate feature flows while preserving the general knowledge learned during pre-training. Adapter methods can be broadly categorized into serial and parallel approaches. Serial adapters \cite{pan2022st, sung2022vl, pfeiffer2020adapterfusion} reshape the output of pre-trained sub-layers, making them suitable for tasks with limited data or requiring high data-fitting capacity. Parallel adapters \cite{Cui_2025_CVPR, Fukuda_2025_CVPR, Yang_2024_CVPR}, on the contrary, integrate alongside pre-trained layers without altering the original forward pass, providing higher stability and better preservation of the generalization ability of the model. 
\section{SECOS Framework}
This section presents the SECOS framework, designed to address the RC-OWSSL task. We first provide a clear definition of RC-OWSSL requirements in \cref{sec:problem}, and then elaborate on the three core modules of SECOS in \cref{sec:GNCSC}, \cref{sec:BWSSC}, and \cref{sec:CAFAS}: Novel Class Semantic Compensation, Batch-Wise Semantic Recapture, and Adapter for Semantic Feature Alignment, respectively.

\subsection{Preliminaries}
\label{sec:problem}
Formally, for an RC-OWSSL task, we define the candidate textual label set $C = C_K \cup C_N$, where $C_K = \{c_1, c_2, \dots, c_k\}$ denotes the known class set, $C_N = \{c_{k+1}, c_{k+2}, \dots, c_{k+n}\}$ denotes the novel class set, and $C_K \cap C_N = \emptyset$. The training set is represented as $D_{train} = D_L \cup D_U$, where $D_L = \{(x_1, y_1), (x_2, y_2), \dots ,(x_l, y_l) \mid y_1, y_2, \dots, y_l \in C_K\}$ is the labeled dataset and $D_U = \{u_1, u_2, \dots, u_m\}$ is the unlabeled dataset. Although samples in $D_U$ have no training labels, each sample inherently corresponds to exactly one class in $C$, implying the absence of out-of-distribution or multi-label instances in both $D_U$ and the test set, denoted as $D_{test}$. The goal of RC-OWSSL is to train a model on $D_{train}$ using an OWSSL algorithm that directly predicts labels from $C$ without Hungarian matching during testing, while optimizing the algorithm for maximal classification performance.

\subsection{Novel Class Semantic Compensation}
\label{sec:GNCSC}
In RC-OWSSL, explicit supervisory signals are heavily imbalanced between known and novel classes. Specifically, $D_L$ contains only labeled data for known classes, while all novel class data are contained within $D_U$. Training a model with $D_{train}$ results in a severe logits bias toward known classes and frequent misclassification of novel class samples. To address this, we first perform Novel Class Semantic Compensation to globally capture latent semantic information from $D_U$, compensating for the lack of supervision in novel classes and balancing it with known classes.

\begin{figure}
\centering
{\includegraphics[width=\columnwidth]{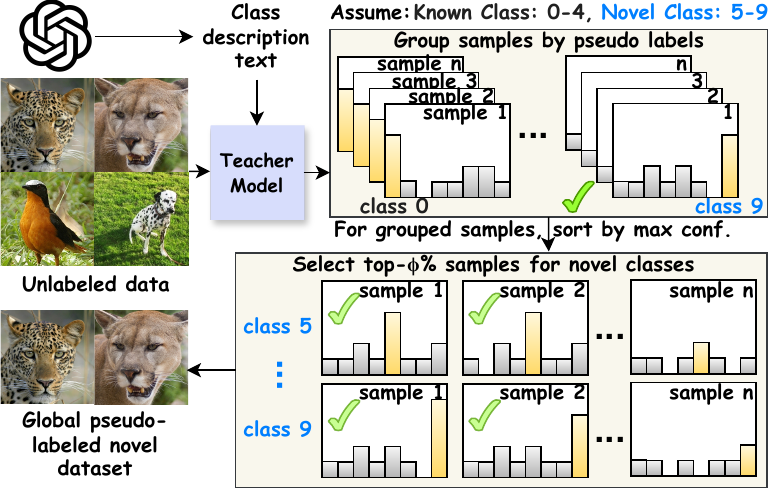}}
\vspace{-15pt}
\caption{Novel Class Semantic Compensation. High-confidence pseudo-labeled novel samples are added to $D_N$ to compensate for the severe lack of supervision in novel classes.}
\label{fig:global_label}
\vskip -15pt
\end{figure}

In particular, as shown in \cref{fig:global_label}, we construct a global pseudo-labeled novel dataset denoted as $D_N$, which stores samples from $D_U$ whose latent semantic information closely corresponds to classes in $C_N$, supplementing the missing supervisory signals for novel classes.

Then, a frozen teacher CLIP model $\mathcal{T}$ is used to assign a hard pseudo label $\hat{y}_i$ to each sample $u_i$ in $D_U$. The pseudo-labeled samples are grouped by $\hat{y}_i$ into $k+n$ subsets, where the subset for class $c \in C$ is $P_c = \{(u_1^c, c), (u_2^c, c), \dots, (u_{m_c}^c, c)\}$ with $m_c$ samples. Only samples with $c \in C_N$ are included in $D_N$. Optionally, the same CLIP backbone can function as both teacher and student through EMA updates, enabling a teacher-free configuration (see \cref{sec:ema}).

Finally, inspired by \cite{liu2025fate, menghini2023enhancing}, for each novel class, the top-$\phi\%$ samples with the highest confidence for class $c$ within $P_c$ 
are added to $D_N$, yielding $D_N = \{(u_1, \hat{y}_1), (u_2, \hat{y}_2), \dots, (u_l, \hat{y}_l) \mid \hat{y}_1, \hat{y}_2, \dots, \hat{y}_l \in C_N\}$. This design ensures that $|D_N|$ is approximately equal to $|D_L|$, achieving globally balanced supervision strength between known and novel classes while maintaining a relatively uniform pseudo-label distribution across $C_N$. Training the model with both $D_N$ and $D_L$ effectively mitigates the strong logits bias toward known classes caused by the absence of supervisory signals for novel classes.

To improve pseudo-label accuracy in $D_N$, we enrich the short textual labels in $C$ with enhanced semantic information. Specifically, a large language model \cite{vaswani2017attention, radford2018improving, brown2020language} generates descriptive prompts for each class, which are encoded by the text encoder of $\mathcal{T}$ and averaged to obtain refined semantic representations. These representations are then used to measure similarity with image features.

\begin{figure}
\centering
{\includegraphics[width=\columnwidth]{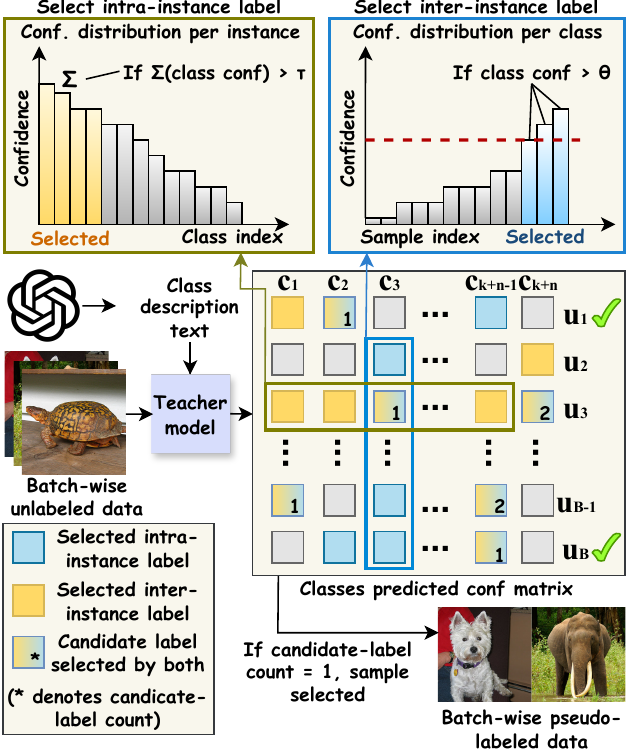}}
\vspace{-15pt}
\caption{Batch-Wise Semantic Recapture. It captures the latent semantic information of each sample at the batch level through both sample-to-class and class-to-sample confidence distributions, further enhancing the supervisory signals in the data.}
\label{fig:local_label}
\vskip -15pt
\end{figure}

\begin{figure*}
\centering
{\includegraphics[width=\textwidth]{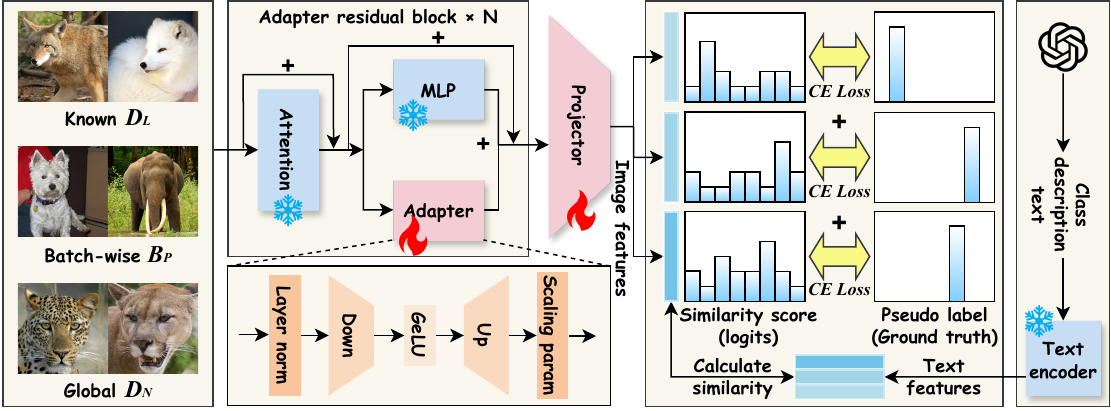}}
\vspace{-20pt}
\caption{Adapter for Semantic Feature Alignment. The adapter aligns visual features with semantic features of textual descriptions prepared in the previous stages, enabling direct prediction of candidate labels for RC-OWSSL.}
\label{fig:adapter}
\vskip -15pt
\end{figure*}

\subsection{Batch-Wise Semantic Recapture}
\label{sec:BWSSC}
Although the Novel Class Semantic Compensation stage effectively compensates substantial semantic information for novel classes, it only performs a coarse-grained extraction, focusing primarily on novel classes. Both neglected semantic information from known classes and unaddressed semantic information from novel classes remain unexploited. To address this, we perform Batch-Wise Semantic Recapture, which captures fine-grained semantic information from $D_U$ to further enrich supervision signals and directly facilitate RC-OWSSL.

Specifically, as shown in \cref{fig:local_label}, $D_U$ is divided into batches, and pseudo labels will be assigned to each batch of unlabeled samples $B = \{u_1, u_2, \dots, u_B\}$. Following \cite{zhang2024candidate}, for each $u_i \in B$, we define an intra-instance label set $C^{intra}_i$ and an inter-instance label set $C^{inter}_i$. $u_i$ is input into $\mathcal{T}$ to obtain its predicted confidence score distribution. Inter-instance labels are added to $C^{intra}_i$ in descending order of their confidence scores for sample $u_i$, until the cumulative confidence just exceeds a threshold $\tau$, where $\tau$ is defined as the $\alpha$-quantile of the maximum confidence scores computed over all samples in $B_U$. For each class $c \in C$, the confidence scores of all samples within $B_U$ for $c$ are sorted in ascending order, and the value at the $\beta$-quantile is selected as the class-specific threshold $\theta_c$. When the confidence score of $u_i$ for $c$ exceeds $\theta_c$, $c$ serves as inter-instance label and added to $C^{inter}_i$. 

For samples with both $C^{intra}_i$ and $C^{inter}_i$, we select $u_i$ where $|C^{intra}_i \cap C^{inter}_i|\!=\!1$ and assign the corresponding candidate label $\hat{y}_i \!=\! C^{intra}_i \cap C^{inter}_i$ to get the batch-wise pseudo-labeled data $B_P \!=\! \{(u_1, \hat{y}_1), (u_2, \hat{y}_2), \dots, (u_b, \hat{y}_b) \mid \hat{y}_1, \hat{y}_2, \dots, \hat{y}_b \in C\}$. This approach recaptures fine-grained semantic information of unlabeled samples from two complementary perspectives: instance-to-class and class-to-instance confidence distributions. It thus selects samples with the most salient and reliable semantic information as pseudo-labeled data within each batch, providing richer and more precise supervisory signals for the RC-OWSSL task.

\subsection{Adapter for Semantic Feature Alignment}
\label{sec:CAFAS}
In the preceding two stages, we have sufficiently captured the latent semantic information embedded within the dataset, providing abundant supervisory signals for each class, particularly the novel classes. Building upon these supervisory signals, the next step is to align the model’s visual feature space with its semantic feature space on downstream data, which enables the model to establish explicit correspondences between each sample and its textual label to complete the final RC-OWSSL task. 

We employ a CLIP \cite{radford2021learning} model as the backbone. Compared with purely visual models \cite{he2016deep, caron2021emerging, dosovitskiy2020image}, CLIP inherently aligns visual and semantic representations during large-scale pre-training, giving them a stronger capability to interpret semantic information in downstream data. We utilize the adapter to achieve multimodal feature alignment on the training set. Concretely, as shown in \cref{fig:adapter}, we freeze the parameters of the text encoder and input the textual descriptions to obtain the semantic feature representation for class $c_i \in C$, denoted as $[E_{c_1}, E_{c_2}, \dots, E_{c_{k+n}}]$. 

For the visual encoder, all original parameters are frozen, and lightweight adapters are inserted after each attention block, in parallel with the subsequent MLP. Each attention block, MLP, and adapter together constitute an adapter residual block, and the visual encoder is composed of $S$ such blocks stacked in series, each with independent parameters. The visual encoder first encodes an input image $x$ to obtain a high-dimensional feature vector, which is then projected through a projector to match the dimensionality of the semantic features. Formally, the process of deriving the high-dimensional visual feature vector $V_x$ of $x$ can be represented as:
\begin{align*}
    &V_x = \text{Projector
    }\bigl(f^{(S)}(x)\bigr), \quad f^{(0)}(x) = x, \\
    &f^{(t+1)}(x) = h^{(t)}+ \text{MLP}(h^{(t)})+ \text{Adapter}(h^{(t)}),
\end{align*}
where $h^{(t)}$ represents the hidden state, denoted as:
\begin{align*}
    h^{(t)} =f^{(t)}(x) + \text{Attention}\bigl(f^{(t)}(x)\bigr).
\end{align*}
Specifically, the adapter is defined as:
\begin{align*}
    \text{Adapter}(z) &= \gamma_{\text{param}} \cdot \mathcal{P}_u\Bigl(\sigma\big(\mathcal{P}_d(\text{LayerNorm}(z))\big)\Bigr),
\end{align*}
where $\gamma_{\text{param}}$ corresponds to the scaling factor in the adapter, and $\mathcal{P}_u$, $\sigma$, $\mathcal{P}_d$ represent the up-sampling projector, the GeLU activation layer, and the down-sampling projector, respectively. We compute the similarity between the visual and textual feature vectors to obtain the predicted logits of the model and calculate the loss of cross-entropy with respect to the ground truth or the hard pseudo label, producing the classification loss for the sample $(x,y)$:
\begin{align*}
    \mathcal{L} (x,y) &= - \sum_{i=1}^{k+n} 
        \mathbb{I}_{[i = y]} 
        \log\Bigl( 
            \exp(\epsilon) \cdot \bigl\langle V_x, E_{c_i} \bigr\rangle 
        \Bigr),
\end{align*}
where $\epsilon$ is the logits scaling factor inherent in the model, and $\langle\cdot,\cdot\rangle$ represents the inner product of two vectors. $y$ also denotes the pseudo label $\hat{y}$ for simplicity, and $(x,y) \in B_L\cup B_N \cup B_P$, where $B_L$, $B_N$ denote the mini-batches corresponding to $D_L$, $D_N$. The final loss is computed as the sum over all three types of labeled data. During training, only the adapter and projector parameters are updated. During inference, all parameters are kept fixed, allowing the model to predict the corresponding textual label for the input image directly.


\section{Experiments} 

\begin{table*}[t]
\caption{Comparison of SECOS with SOTA OWSSL and GCD methods on generic datasets. \textbf{Bold} and \underline{underline} indicate the best and second-best results, respectively. K/N/A denotes Known, Novel, and All accuracy, respectively. }
\label{tab:generic}
\vskip -5pt
\centering
\setlength{\tabcolsep}{8pt}
\begin{tabular}{l|c|ccc|ccc|ccc}
\toprule
 & &\multicolumn{3}{c|}{CIFAR10} & \multicolumn{3}{c|}{CIFAR100} & \multicolumn{3}{c}{ImageNet100} \\
Methods& Backbone & K & N & A & K & N & A & K & N & A \\
\midrule
TIDA \cite{wang2023discover} & ResNet &94.2&93.4&93.8&73.3&56.6&65.3&83.4&71.2&77.6\\
OwMatch\cite{niu2024owmatch} & ResNet &96.5&97.1&96.8&80.1&63.9&71.9&91.5&79.6&85.5\\
TRAILER \cite{xiao2024targeted} & ResNet &93.4&95.0&94.4&69.7&48.7&55.6&91.4&82.4&85.3\\
GCD \cite{vaze2022generalized}  & DINO   & 97.9 & 88.2 & 91.5 & 76.2 & 66.5 & 73.0 & 89.8 & 66.3 & 74.1 \\
SimGCD \cite{wen2023parametric} & DINO & 95.1 & 98.1 & 97.1 & 81.2 & 77.8 & 80.1 & 93.1 & 77.9 & 83.0 \\
DCCL \cite{pu2023dynamic} & DINO & 96.5 & 96.9 & 96.3 & 76.8 & 70.2 & 75.3 & 90.5 & 76.2 & 80.5 \\
GPC \cite{zhao2023learning} & DINO  & \textbf{98.2} & 89.1 & 92.2 & {85.0} & 63.0 & 77.9 & \underline{94.3} & 71.0 & 76.9 \\
PromptCAL \cite{zhang2023promptcal} & DINO & 96.6 & 98.5 & 97.9 & 84.2 & 75.3 & 81.2 & 92.7 & 78.3 & 83.1 \\
SPTNet  \cite{wang2024sptnet} & DINO & 95.0 & \underline{98.6} & 97.3 & 84.3 & 75.6 & 81.3 & 93.2 & 81.4 & 85.4 \\
SimGCD \cite{wen2023parametric} & CLIP & 94.7 & 97.5 & 96.6 & 82.6 & \underline{79.5} & \underline{81.6} & \textbf{94.5} & 81.9 & 86.1 \\
TextGCD \cite{zheng2024textual} & CLIP& \underline{98.0} & \underline{98.6} & \textbf{98.2} & \underline{85.5} & 71.6 & 80.8 & 92.4 & \underline{85.2} & \underline{88.0} \\
TP-OWSSL \cite{fan2025learning}& CLIP & 92.4 & 95.4 & 94.4 & 71.5 & 72.9 & 72.3 & 84.2 & 86.8 & 85.8 \\
\midrule
SECOS (ours)   & CLIP      & 97.9 & \textbf{98.8} & \underline{98.1} & \textbf{87.1} & \textbf{82.6} & \textbf{84.7} & 93.4 & \textbf{89.8} & \textbf{91.7} \\
\bottomrule
\end{tabular}
\vskip -5pt
\end{table*}

\begin{table*}[t]
\caption{Comparison of SECOS with SOTA OWSSL and GCD methods on fine-grained datasets.}
\label{tab:fine-grained}
\vskip -5pt
\centering
\begin{tabular}{l|c|ccc|ccc|ccc|ccc}
\toprule
 &  &\multicolumn{3}{c|}{CUB} & \multicolumn{3}{c|}{Stanford Cars} & \multicolumn{3}{c|}{Oxford Flowers}& \multicolumn{3}{c}{Oxford Pets} \\
Methods&Backbone &K & N & A & K & N & A & K & N & A& K & N & A\\
\midrule
GCD  &DINO   & 56.6          & 48.7          & 51.3          & 57.6          & 29.9          & 39.0          & 74.9          & 74.1          & 74.4          & 85.1          & 77.6          & 80.2                \\
SimGCD & DINO   &65.6          & 57.7          & 60.3          & 71.9          & 45.0          & 53.8          & 80.9          & 66.5          & 71.3          & 85.9          & 88.6          & 87.7              \\
SimGCD & CLIP & 76.8          & 54.6          & 62.0          & 81.4          & 73.1          & 75.9          & 87.8          & 69.0          & 75.3          & 75.2          & \underline{95.7}  & 88.6                 \\
TextGCD & CLIP & 80.6          & \underline{74.7}          & \underline{76.6}        & 87.4          & 86.8          & 86.9          & 90.7          & \underline{85.4}  & 87.2          & 93.9          & \textbf{96.4} & \textbf{95.5}      \\
TP-OWSSL & CLIP & \textbf{83.7} & 72.6 & 76.6 & \textbf{95.5} & \underline{88.6}  & \underline{90.7}  & \underline{90.9}  & \textbf{86.9} & \underline{87.3}  & \underline{94.7}          & 93.5          & 93.9         \\
\midrule
SECOS (ours)        &         CLIP            & \underline{82.0}  & \textbf{75.1}  & \textbf{78.6}  & \underline{89.8}  & \textbf{94.4} & \textbf{92.3} & \textbf{98.4} & 83.7          & \textbf{90.0} & \textbf{95.3} & 95.5          & \underline{95.4}   \\

\bottomrule
\end{tabular}
\vskip -10pt
\end{table*}

\subsection{Basic Settings} 
\textbf{Dataset.} We evaluate SECOS on three generic image classification datasets: CIFAR10 \cite{krizhevsky2009learning}, CIFAR100 \cite{krizhevsky2009learning}, and ImagenNet100 \cite{krizhevsky2012imagenet}, as well as four fine-grained datasets: CUB \cite{wah2011caltech}, Stanford Cars \cite{krause20133d}, Oxford Flowers \cite{nilsback2008automated}, and Oxford Pets \cite{parkhi2012cats}. For each dataset, we designate the first half of the classes as known classes and the remaining half as novel classes. Among the training samples belonging to the known classes, 50\% are selected to form the labeled dataset. The remaining samples from the known classes, together with all samples from the novel classes, are used to construct the unlabeled dataset. For datasets without predefined train-test splits, including Flowers, Cars, and Pets, we use the entire dataset for training and randomly sample 20\% of unlabeled data as the balanced test set. \\
\textbf{Evaluation Protocol.} For all comparison methods \cite{wang2023discover, niu2024owmatch, xiao2024targeted, vaze2022generalized, wen2023parametric, pu2023dynamic, zhao2023learning, zhang2023promptcal, wang2024sptnet, zheng2024textual, fan2025learning}, unless otherwise specified, we follow their original testing protocol. This protocol uses Hungarian matching to compute clustering accuracy, formulated as $ACC_{cluster}\!=\!\dfrac{1}{|D_{test}|}\sum_{i\!=\!1}^{|D_{test}|}\mathbb{I}\big(y_i\!=\!W(\bar{y}_i)\big)$, where $\bar{y}$ is the predicted label and $W$ denotes the optimal permutation. In contrast, SECOS computes the classification accuracy by comparing the predicted labels directly with the ground truth, formulated as $ACC_{classify}\!=\!\dfrac{1}{|D_{test}|}\sum_{i\!=\!1}^{|D_{test}|}\mathbb{I}(y_i\!=\!\bar{y}_i)$. It is worth noting that for the same algorithm, $ACC_{cluster}\!\ge\!ACC_{classify}$, since Hungarian matching reassigns predicted labels to maximize overall accuracy regardless of semantic consistency.\\
\textbf{Implementation Details.} We follow prior works \cite{zheng2024textual, fan2025learning} and employ the same teacher models and backbones: CLIP-ViT-H-14, released by OpenCLIP, and CLIP-ViT-B-16, released by OpenAI. The training process is conducted for 100 epochs with a batch size of 32, using the AdamW optimizer with an initial learning rate of 0.0001, weight decay of 1e-5, and a LambdaLR scheduler. Following \cite{zhang2024candidate}, we set the hyperparameters $\alpha\!=\!0.6$ and $\beta\!=\!0.95$ in the Batch-Wise Semantic Capture stage. The global pseudo-labeled novel samples percentage $\phi$ is set to 50, and the downsampling dimension of the adapter is set to 10, which will be discussed in detail in \cref{sec:novel_count} and \cref{sec:low_dim}, respectively. During loss computation, each image is augmented into a strong view and a weak view \cite{sohn2020fixmatch}. The classification losses of the two views are calculated and summed to obtain the final loss for the image. All experiments are conducted using PyTorch \cite{paszke2019pytorch} with one GeForce RTX 3090 GPU.

\subsection{Main Results} 
We compare SECOS with existing OWSSL and GCD SOTA methods. \cref{tab:generic} reports the results on generic datasets, while \cref{tab:fine-grained} presents those on fine-grained datasets. As shown, even though previous methods evaluate their clustering accuracy for Novel and All using Hungarian matching. At the same time, SECOS adheres to the standard classification accuracy, and SECOS still achieves significantly superior performance. In particular, SECOS achieves the highest performance across all three generic datasets in Novel accuracy and performs best on two datasets and second-best on one in All accuracy. On the fine-grained datasets, SECOS attains two best results in Novel accuracy, as well as three best and one second-best results in All accuracy. Compared with TP-OWSSL \cite{fan2025learning}, a framework tailored for fine-grained OWSSL classification, SECOS still achieves an improvement of 1.78\% and  1.95\% in Novel and All accuracy on fine-grained datasets, and 7.37\% and 5.35\% improvements on generic datasets, respectively. These results demonstrate that SECOS effectively handles downstream datasets with either large or small inter-class distribution gaps, exhibiting strong robustness and generalization capability.

\begin{table}
\caption{Ablation study of the two semantic capture stages in SECOS on CIFAR100 and CUB datasets. ``N” and ``B” denote the Novel Class Semantic Compensation and Batch-Wise Semantic Recapture stages, respectively.}
\label{tab:ablation}
\vskip -5pt
\centering
\setlength{\tabcolsep}{6pt}
\begin{tabular}{cc|ccc|ccc}

\toprule
\multicolumn{2}{c|}{Ablation} & \multicolumn{3}{c|}{CIFAR100}& \multicolumn{3}{c}{CUB}\\
 N & B      & K         & N         & A          & K          & N         & A  \\
\midrule
 \ding{55} & \ding{55} & \textbf{90.0} & 14.4          & 50.7        & 79.5          & 22.6          & 51.2  \\
\ding{51} & \ding{55}   & 86.7          & \uline{80.4}  & \uline{83.9}  & \uline{79.6}          & \uline{74.5}  & \uline{77.5} \\
  \ding{55} & \ding{51}  & \uline{87.8}  & 76.7          & 82.3          & \textbf{82.0}  & 69.1          & 74.6    \\
  \ding{51}  &  \ding{51}  & 87.1          & \textbf{82.6} & \textbf{84.7} & \textbf{82.0} & \textbf{75.1} & \textbf{78.6} \\
\bottomrule
\end{tabular}
\vskip -20pt
\end{table}

\begin{figure*}
\centering
{\includegraphics[width=\textwidth]{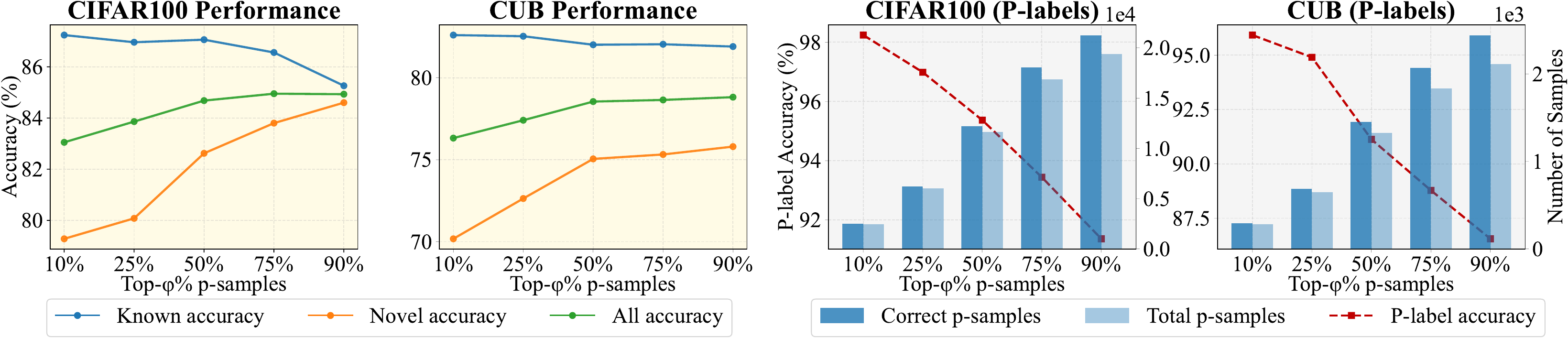}}
\vspace{-20pt}
\caption{Analysis of global novel pseudo-label (p-label) density. The left two subfigures show the variation of Known, Novel, and All accuracy with respect to the proportion of top-$\phi$\% pseudo-labeled samples (p-samples). The right two subfigures present the corresponding pseudo-label accuracy, total number of pseudo-labeled samples, and number of correct pseudo-labeled samples as $\phi$\ increases.}
\label{fig:global_novel_count}
\vskip -15pt
\end{figure*}

\begin{figure} 
\centering
{\includegraphics[width=\columnwidth]{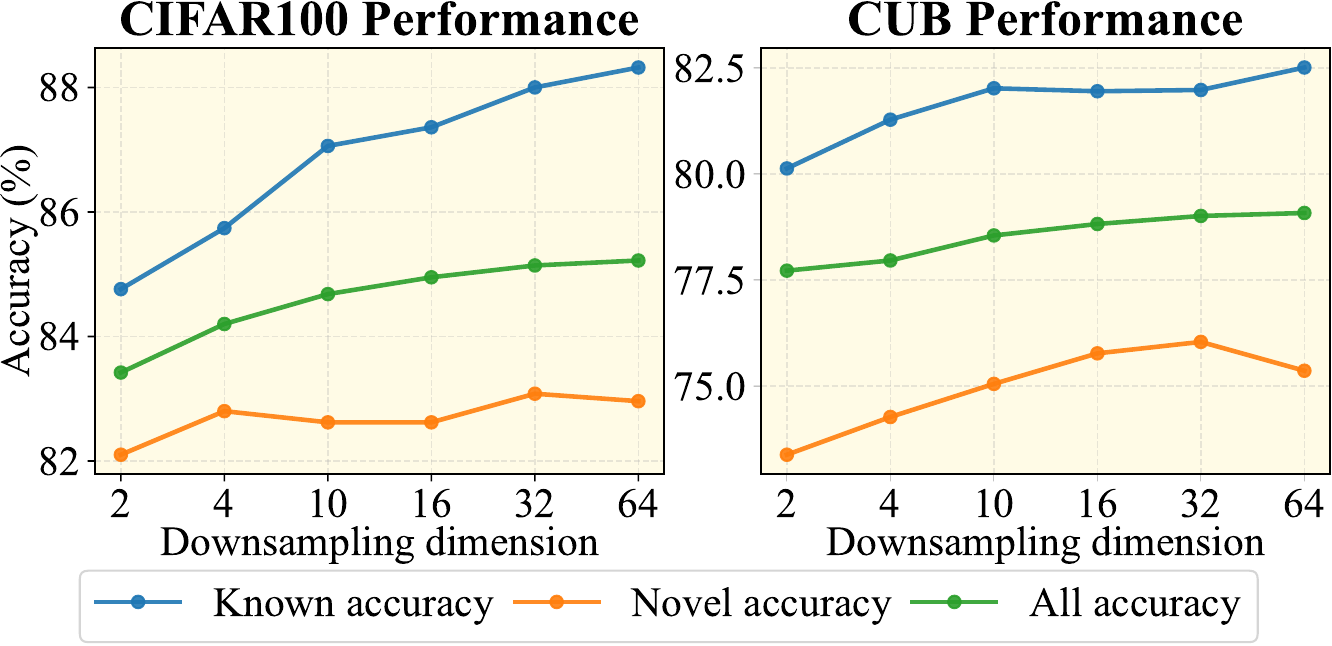}}
\vspace{-15pt}
\caption{Effect of the adapter's downsampling dimension on model performance. Increasing the downsampling dimension from 2 to 64 yields less than a 2\% improvement in All accuracy.}
\label{fig:low_dim}
\vskip -15pt
\end{figure}

\subsection{Ablation Study} 
To validate the necessity of the two semantic capture stages in SECOS, we conduct ablation studies on the CIFAR100 and CUB datasets. Using the model trained solely on $D_L$ as the baseline, we progressively incorporate the two semantic capture modules, and the results are presented in \cref{tab:ablation}. As shown, when trained only with $D_L$, the model exhibits a severe logits bias toward known classes due to the significant imbalance in supervisory signals between known and novel classes, resulting in extremely low Novel accuracy. After introducing the two semantic capture strategies, Novel accuracy improves substantially. Specifically, the Novel Class Semantic Compensation stage globally supplements the supervisory signals of novel classes. In contrast, the Batch-Wise Semantic Recapture stage further captures fine-grained semantic information available within the dataset. Consequently, incorporating the global stage leads to a pronounced improvement in Novel accuracy, and adding the batch-wise stage further enhances All accuracy.



\subsection{Novel Pseudo-Label Density Analysis}
\label{sec:novel_count}
We examine how the number of global pseudo-labeled novel samples influences model performance. Specifically, we set $\phi$ to 10, 25, 50 (approximately matching the number of labeled samples), 75, and 90 to construct $D_N$ for training. The results are shown in \cref{fig:global_novel_count}. When only a small number of high-confidence samples are used, both Novel and All accuracy are relatively low. This occurs because the supervision for novel classes remains considerably weaker than that for known classes, even though the pseudo labels are more accurate, resulting in a pronounced training bias toward known classes. As the proportion of pseudo-labeled samples increases, Known accuracy gradually decreases while Novel accuracy improves, and All accuracy tends to stabilize. Similarly, as more pseudo-labeled samples are incorporated, noise is inevitably introduced. When the number of pseudo-labeled novel samples surpasses that of labeled known samples, a trade-off between accuracy and noise emerges, resulting in an overall stable performance.

\subsection{Influence of Adapter Downsampling Dimension} 
\label{sec:low_dim}
To investigate the impact of the adapter’s downsampling dimension on the performance of SECOS, we conduct experiments on the CIFAR100 and CUB datasets, setting the adapter’s downsampling dimension to 2, 4, 10, 16, 32, and 64. The results are illustrated in \cref{fig:low_dim}. It is shown that as the downsampling dimension increases, the All accuracy on both datasets gradually improves but quickly converges, with a total gain of less than 2\% between the smallest and largest dimensions. This indicates that the adapter’s downsampling dimension has a minimal effect on the overall performance of SECOS, demonstrating its strong stability. Considering the trade-off between performance and resource consumption during training, it is generally unnecessary to set an enormous low-dimensional size (e.g., 64 or higher); a moderate value, such as 10, is sufficient for maintaining high performance while ensuring efficiency.

\begin{table}
\caption{Comparison of model performance trained on pseudo-labeled samples (p-samples) obtained from two candidate label selection strategies. }
\label{tab:batch-size}
\vskip -5pt
\centering
\setlength{\tabcolsep}{3.875pt}
\begin{tabular}{l|ccc|ccc}
\toprule
& \multicolumn{3}{c|}{CIFAR100}& \multicolumn{3}{c}{CUB}\\
Training on       & K         & N         & A          & K          & N         & A  \\
\midrule
original p-samples & 85.2    & 80.6 & 82.4 & 78.5 & 70.7 & 75.3\\
filtered p-samples & \textbf{87.1}    & \textbf{82.6} & \textbf{84.7} &\textbf{ 82.0} & \textbf{75.1} & \textbf{78.6}\\
\bottomrule
\end{tabular}
\vskip -10pt
\end{table}

\begin{figure*}  
\centering  
{\includegraphics[width=\textwidth]{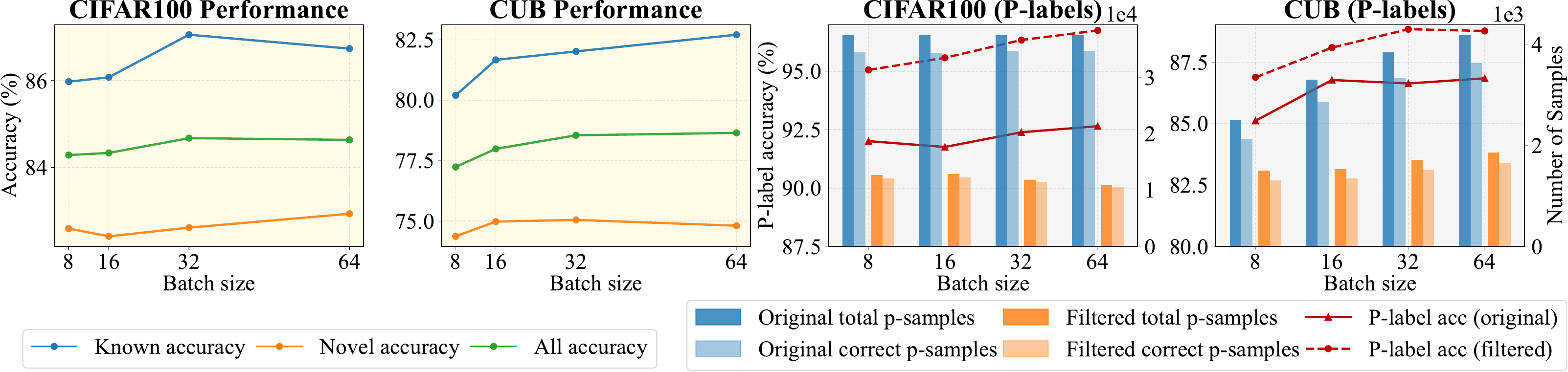}}
\vspace{-20pt}
\caption{Impact of batch size and pseudo-label selection on SECOS. The left two figures show that model performance remains stable across batch sizes. The right two figures show pseudo-label precision and the number of pseudo-labeled samples for two candidate label selection strategies, indicating that retaining samples with exactly one pseudo label consistently yields higher precision.}
\label{fig:batch_size}
\end{figure*}

\begin{table*}

\caption{Performance comparison between SECOS with and without teacher supervision. The SECOS (w.o. teacher) variant shows only a minor drop in overall accuracy while still outperforming existing OWSSL methods, demonstrating its strong self-learning capability.}
\label{tab:ema}
\centering
\vskip -5pt
\begin{tabular}{l|c|ccc|ccc|ccc|ccc}
\toprule
& &\multicolumn{3}{c|}{CIFAR100}   & \multicolumn{3}{c|}{ImageNet100}& \multicolumn{3}{c|}{Oxford Flowers}&\multicolumn{3}{c}{Oxford Pets}\\
               Method  & Teacher model    & K              & N              & A              & K              & N              & A              & K              & N              & A              & K              & N              & A      \\
\midrule
TextGCD &  \ding{51}  & \uline{85.5}  & 71.6  & 80.8  & \uline{92.4}  & 85.2  & 88.0  & 90.7  & \uline{85.4}  & 87.2  & 93.9  & \uline{96.4}  & \textbf{95.5} \\
TP-OWSSL &  \ding{51}  & 71.5  & 72.9  & 72.3  & 84.2  & 86.8  & 85.8  & 90.9  & \textbf{86.9}  & 87.3  & \uline{94.7}  & 93.5  & 93.9 \\
\midrule
SECOS& \ding{51} & \textbf{87.1}  & \textbf{82.6}  & \textbf{84.7}  & \textbf{93.4}  & \uline{89.8}  & \textbf{91.7}  & \textbf{98.4}  & 83.7  & \textbf{90.0}  & \textbf{95.3}  & 95.5  & \uline{95.4} \\
SECOS& \ding{55} & 84.3  & \uline{81.1}  & \uline{82.4}  & 91.9  & \textbf{90.7}  & \uline{91.3}  & \uline{96.8}  & 82.0  & \uline{88.3}  & 94.5  & \textbf{96.5}  & \textbf{95.5} \\
\bottomrule
\end{tabular}
\vskip -10pt
\end{table*}

\subsection{Batch-Wise Pseudo-Labeling Strategy Analysis}
We investigate the impact of batch size and pseudo-label selection strategy on the batch-wise pseudo-labeling process. We present experiments with batch sizes of 8, 16, 32, and 64. As shown in \cref{fig:batch_size} (left), the performance degradation across all three metrics is minimal between the smallest and largest batch sizes, with only a 0.35\% drop in overall accuracy on CIFAR100 and a 1.41\% drop on CUB. This demonstrates that SECOS is robust to variations in batch size.

We further examine the influence of pseudo-label precision by comparing two candidate label selection strategies: one using the raw intersection of intra- and inter-instance label sets, and another retaining only samples with one pseudo label in the candidate label set. As shown in \cref{fig:batch_size} (right), both strategies show slight improvements as the batch size increases, yet the latter consistently achieves much higher pseudo-label precision. This is because when the candidate label set contains multiple labels, the semantic representation of the sample is ambiguous and easily confused with other classes, leading to lower pseudo-label accuracy. In contrast, when the candidate label set contains exactly one label, the sample’s semantic information is well-defined and strongly indicative of its true class, resulting in significantly improved pseudo-label quality.

Subsequently, we use the pseudo-labeled samples obtained from these two strategies to train the model. As shown in \cref{tab:batch-size}, the model trained on samples with one pseudo label in the candidate label set achieves better performance. Although the number of filtered pseudo-labeled samples is much smaller than that of the original pseudo-labeled samples, as illustrated in \cref{fig:batch_size} (right), this result confirms that clearer semantic assignments lead to higher-quality pseudo labels and more effective learning.

\subsection{Self-Learning Capability of SECOS}
\label{sec:ema}
SECOS exhibits robust self-learning ability and can perform effectively on some tasks even without relying on a teacher model. We directly use the student CLIP to generate pseudo labels during the Novel Class Semantic Compensation stage. At each parameter update, an EMA-based copy of the student CLIP is maintained and used to produce pseudo labels during the Batch-Wise Semantic Recapture stage, while other processes of SECOS remain unchanged. The experimental results, presented in \cref{tab:ema}, show that compared with original SECOS, SECOS without a teacher model achieves only a 0.83\% average decrease in overall performance across four datasets, while still outperforming existing OWSSL methods that rely on teacher models. This demonstrates that SECOS can effectively adapt to scenarios without teacher supervision, validating the strength of its design in fully exploiting CLIP to capture latent semantic information and enhance training supervision.
\section{Conclusion}
This paper introduces the RC-OWSSL task, where existing OWSSL methods fall short due to insufficient semantic grounding between visual and textual modalities, emphasizing direct textual label prediction without post-hoc matching. To overcome this limitation, SECOS is proposed, leveraging external knowledge to extract and align semantic representations from both known and novel classes, enabling direct prediction of textual labels. Extensive experiments across multiple benchmarks show that SECOS consistently surpasses previous SOTA methods, even those evaluated with post-hoc alignment, highlighting its effectiveness in achieving practical and rigorous open-world classification.\\
\noindent\textbf{Limitations.} SECOS depends on large-scale vision-language models such as CLIP, and its performance is inherently tied to the quality and coverage of the underlying image-text pretraining. However, given that RC-OWSSL fundamentally requires semantic grounding across modalities, such dependence is natural and shared among recent comparable methods.\\
\clearpage                

\section*{Acknowledgment}
This study was supported in part by the National Natural Science Foundation of China under Grants 62431004, 62376233, 62502402, 62476063 and 62306181; in part by the Natural Science Foundation of Fujian Province under Grant 2024J09001; in part by the Natural Science Foundation of Guangdong Province under Grant 2025A1515011293; in part by the Guangdong Basic and Applied Basic Research Foundation under Grant 2024A1515010163; in part by the Shenzhen Science and Technology Program under Grant RCBS20231211090659101; in part by the National Key Laboratory of Radar Signal Processing under Grant JKW202403; in part by Xiaomi Young Talents Program; and SNG is supported by the Computational Science Centre for Research Communities (CoSeC) Fellowship programme.

{
    \small
    \bibliographystyle{ieeenat_fullname}
    \bibliography{main}
}


\end{document}